\documentclass{article}

\usepackage{amsmath}
\usepackage{comment}
\usepackage{graphicx}
\usepackage{subfig}

\usepackage[preprint]{neurips_2024}

\usepackage[utf8]{inputenc} %
\usepackage[T1]{fontenc}    %
\usepackage{hyperref}       %
\usepackage{url}            %
\usepackage{booktabs}       %
\usepackage{amsfonts}       %
\usepackage{nicefrac}       %
\usepackage{microtype}      %
\usepackage{xcolor}         %
\usepackage{algorithm}
\usepackage{algorithmicx}
\usepackage{algpseudocode}
\usepackage{listings}
\usepackage{wrapfig}
\title{Tangent Space Fine-Tuning for Directional Preference Alignment in Large Language Models}

\author{%
  Mete Erdogan \\
  Department of Electrical Engineering \\
  Stanford University \\
  \texttt{merdogan@stanford.edu} \\
  \vspace{0cm} \\
  \texttt{GitHub:} \url{https://github.com/meterdogan07/TSDPO}
}

\begin{document}

\maketitle

\begin{abstract}
Our goal is to enable large language models (LLMs) to balance multiple human preference dimensions; such as helpfulness, safety, and verbosity, through principled and controllable alignment. Existing preference optimization methods, including Direct Preference Optimization (DPO), collapse feedback into a single scalar reward, fixing one balance among objectives and preventing traversal of the Pareto front. Recent work by \citet{ortiz2023task} showed that fine-tuning can be viewed in a model’s \emph{tangent space}, where linearized updates act as additive vectors that can be composed to jointly perform well on multiple tasks. Building on this formulation, we extend this idea to preference alignment and propose \emph{Tangent-Space Direct Preference Optimization (TS-DPO)}, which performs DPO within this locally linear regime to learn per-objective update directions. These directions can be linearly combined at inference (e.g., $\theta=\theta_0+\alpha\,\tau_{\text{helpful}}+\beta\,\tau_{\text{verbose}}$) to generate user-specified behaviors without additional optimization. Evaluated on the helpfulness–verbosity trade-off using the HelpSteer and UltraFeedback datasets, TS-DPO achieves broader Pareto-optimal coverage and smoother preference control than scalarized DPO. Canonical Correlation Analysis (CCA) further shows that tangent-space training amplifies canonical directions aligned with distinct preferences, improving disentanglement. 
\end{abstract}

\section{Introduction}
\vspace{-0.2cm}

As large language models (LLMs) increasingly operate in personalized and safety-critical settings, their responses are evaluated along \textit{multiple} human preference dimensions such as helpfulness, safety, and verbosity. These dimensions often conflict: a highly detailed response may violate a user’s preference for concise communication, whereas an overly terse answer may sacrifice clarity. This motivates the formulation of alignment as a \emph{multi-objective preference optimization} problem rooted in human choice behavior. 

Let $\theta \in \mathbb{R}^d$ denote model parameters and 
\[
\mathbf{f}(x;\theta) = (f_1(x;\theta), \dots, f_K(x;\theta))
\]
represent latent utilities over $K$ objectives. Pairwise judgments impose constraints of the form
$(x^+ \succ x^-) \Rightarrow f_k(x^+;\theta) \ge f_k(x^-;\theta)$,
consistent with the Bradley–Terry model. We say $\mathbf{f}(x;\theta') \succeq \mathbf{f}(x;\theta)$ if $f_k(x;\theta') \ge f_k(x;\theta)$ for all $k$. The \emph{Pareto-optimal set} is
\[
\Theta^\star = \{\theta ~|~ \not\exists\, \theta' : \mathbf{f}(x;\theta') \succeq \mathbf{f}(x;\theta) \text{ and } \mathbf{f}(x;\theta') \neq \mathbf{f}(x;\theta)\},
\]
where each $\theta \in \Theta^\star$ encodes a distinct trade-off among objectives.

Direct Preference Optimization (DPO) combines all reward components into one scalar objective, so the optimization process converges to a single trade-off among competing preferences. This yields only a single point on $\Theta^\star$. Therefore, switching to a different preference configuration requires re-training, because DPO does not learn the underlying preference directions separately. The key question of this work is therefore:
\begin{quote}
\emph{Can we learn disentangled and composable preference directions by applying DPO in the tangent space of a pre-trained model?}
\end{quote}

Our hypothesis is that linearizing the model around its base parameters enables each preference dimension (e.g., helpfulness, verbosity) to correspond to an independent update vector, allowing smooth interpolation among Pareto-optimal trade-offs at inference. Modular preference alignment is crucial for adaptable, value-sensitive AI systems. TS-DPO could enable LLMs to adjust behavior dynamically across users or contexts, contributing to safer, practical, and more interpretable alignment methods. \looseness=-2

This viewpoint reframes alignment as learning additive directions in parameter space rather than optimizing a single monolithic set of weights. Such modular preference alignment would enable LLMs to dynamically adjust behavior across users, applications, or deployment contexts. \looseness=-2

Beyond performance evaluation, we analyze the representational effects of TS-DPO by comparing how training modifies the model’s internal geometry relative to the base model. Using canonical correlation analysis (CCA) and activation-subspace overlap, we test whether preference-specific updates align with distinct canonical directions in representation space. This allows us to examine not only whether TS-DPO matches or surpasses conventional DPO in performance, but also whether it induces more interpretable and disentangled changes in the model’s internal computations.

\vspace{-0.2cm}
\section{Literature Review}
\vspace{-0.2cm}
Direct Preference Optimization (DPO) \citep{}{dpo2023} optimizes model parameters directly from human pairwise comparisons without learning an explicit reward model. Given a preferred response $x^+$ and a dispreferred response $x^-$ to the same prompt $x$, DPO maximizes the log-likelihood margin between the two under a Bradley–Terry assumption:
\begin{equation}
\mathcal{L}_{\text{DPO}}(\pi_\theta; \pi_{\text{ref}})
= - \mathbb{E}_{(x,y_w,y_l) \sim \mathcal{D}} \Big[
\log \sigma \big(
\beta \log \tfrac{\pi_\theta(y_w \mid x)}{\pi_{\text{ref}}(y_w \mid x)}
- \beta \log \tfrac{\pi_\theta(y_l \mid x)}{\pi_{\text{ref}}(y_l \mid x)}
\big)
\Big],
\label{eq:dpo_loss}
\end{equation}
where $\pi_{\text{ref}}$ is a static reference model and $\beta$ controls conservativeness. While effective and scalable, DPO implicitly assumes a single scalar reward and therefore learns only one operating point in the multi-objective utility space, limiting control over trade-offs among preferences.

SteerLM \citep{dong2023steerlm} proposes a multi-attribute alignment framework using attribute classifiers that condition generation on explicit dimensions (e.g., helpfulness, humor, toxicity). SteerLM jointly predicts human annotations and conditions generation on these attributes, enabling smooth preference control without reward modeling or RL. This demonstrates that multi-attribute alignment is feasible without collapsing objectives into a single scalar, further motivating compositional control.

Directional Preference Alignment (DPA) \citep{dpa2024} tackles the same problem by learning a vector-valued reward model $\mathbf{r}(x,y)\in\mathbb{R}^K$ for different preference axes. Each user-specified direction $v\in S_K$ defines a scalarized reward $R(x,v,y)=v^\top \mathbf{r}(x,y)$, and the conditional policy $\pi_\theta(y\!\mid\!x,v)$ is optimized to maximize $\mathbb{E}[R(x,v,y)]$ via rejection-sampling fine-tuning \citep{dong2023raft}. At inference, the same model can be steered along the Pareto front by varying $v$. However, because DPA performs scalarized training, the learned solutions do not compose linearly in parameter space, restricting modularity across objectives.

A separate line of work has shown that models fine-tuned on different objectives can be merged directly in weight space, by parameter interpolation and model merging strategies. \cite{wortsman2022model} formalized these observations, such that independently fine-tuned models can be averaged in parameter space to improve accuracy and approximate ensembles, suggesting that weight updates behave in a surprisingly linear fashion. This points to the existence of low-curvature regions in the loss landscape where parameter directions can be combined meaningfully.

Building on these observations, a recent work on \emph{task vectors} \citep{ortiz2023task} shows that fine-tuning updates behave approximately linearly in the model’s \emph{tangent space}, defined by the first-order expansion:
\[
f(x;\theta)\approx f(x;\theta_0)+J_{\theta_0}(x)(\theta-\theta_0),
\]
where $J_{\theta_0}$ is the Jacobian at initialization. The resulting update $\tau=\theta-\theta_0$ acts as an additive direction in parameter space whose semantic effect can combine with others. This motivates our exploration of whether distinct preference directions can be represented as linear, composable updates under the DPO framework. \looseness=-2

To our knowledge, tangent-space decomposition has not yet been investigated in the context of multi-objective preference optimization. Our study aims to explore whether combining these ideas yields practical benefits. \looseness=-2

Finally, \cite{raghu2017svcca} introduced \emph{Singular Vector Canonical Correlation Analysis (SVCCA)} to compare neural representations across layers and models in various pre-training phases. SVCCA identifies correlated subspaces that capture most representational variance, revealing shared low-dimensional structure. Inspired by this, we employ CCA to study the geometry of our tangent-space preference updates to assess whether multi-objective alignment directions remain disentangled, composable, and interpretable. \looseness=-2

\vspace{-0.2cm}
\section{Proposed Method: Tangent-Space Direct Preference Optimization}
\vspace{-0.2cm}
We propose Tangent-Space Direct Preference Optimization (TS-DPO), a preference learning  framework in which different alignment objectives are trained as separate update directions around a fixed base model using DPO. The method relies on modeling the fine-tuning as a first-order approximation around the base parameters, so that changes in model behavior can be written as the dot-product between parameter updates and the Jacobian at the initialization point. As a result, each objective corresponds to an additive direction in parameter space, and the final aligned model can be expressed as the base model plus a weighted combination of objective-specific directions. The coefficients controlling each direction become user-selected knobs that adjust behavior at inference-time.

\vspace{-0.2cm}
\subsection{Linearization of Parameter Space and Linearized Model Construction}
\vspace{-0.2cm}
TS-DPO operates in the tangent space of a frozen pretrained model, instead of optimizing the entire parameter vector. Let $\theta_{0}$ denote the parameters of the base instruction-tuned model. Using a first-order Taylor expansion of the model around $\theta_0$, the output under an update $\Delta\theta$ is approximated by:
\begin{equation}
f(x;\theta_0 + \Delta\theta)
\approx
f(x;\theta_0) + J_{\theta_0}(x)\,\Delta\theta,
\label{eq:linearized_model}
\end{equation}
where $J_{\theta_{0}}(x)$ is the Jacobian of model's output with respect to parameters. This shows that preference learning can be expressed entirely in terms of the update direction $\Delta\theta$ when base model is frozen. \looseness=-2

Following the task-vector formulation of \citet{ortiz2023task}, we represent alignment as learning additive update directions in parameter space. In contrast to the implementation of \citet{ortiz2023task}, where the updated model parameters $\theta_{0} + \Delta\theta$ are typically stored or combined, we maintain the tangent-space updates $\Delta\theta$ directly as a trainable set of parameters \texttt{dparams}. \looseness=-2

To construct this linearized model, the pretrained network is converted to functional form using \texttt{make\_functional\_with\_buffers} function of the \texttt{functorch}\footnote{\url{https://docs.pytorch.org/functorch/stable/}} library, to extract the initial parameters $\theta_0$ as a constant reference point. A corresponding set of trainable variables \texttt{dparams} is introduced to represent the tangent update direction. Thus, instead of materializing the full updated parameter vector, we keep the frozen base parameters and the learned update directions separate. \looseness=-2

The forward computation then combines the frozen model output with the effect of the update direction using a single Jacobian–vector product (JVP), as shown in Algorithm~\ref{alg:linear}. Importantly, the base parameters $\theta_0$ are never modified; only \texttt{dparams} are optimized. This keeps TS-DPO strictly first-order with respect to the initial model. Formally, the forward pass computes the Equation \ref{eq:linearized_model}, where the JVP operator evaluates the tangent contribution efficiently without ever materializing the full Jacobian. \looseness=-2

\begin{algorithm}[t]
\caption{Linearized model with frozen base parameters}
\setlength{\belowcaptionskip}{-5pt}
\begin{lstlisting}[
    language=Python,
    basicstyle=\ttfamily\scriptsize,
    keywordstyle=\color{blue}\bfseries,
    stringstyle=\color{orange},
    commentstyle=\color{teal},
    frame=single,
    showstringspaces=false,
    columns=fullflexible,
    morekeywords={nn, jvp, make_functional_with_buffers}
]
from functorch import jvp, make_functional_with_buffers

class LinearizedModel(nn.Module):
    """
    Linearized model: f(x; params0 + dparams)
    (params0 frozen; only dparams are updated)
    """
    def __init__(self, init_model):
        func, params0, buffers0 = make_functional_with_buffers(init_model)

        # store frozen params
        self.params0 = params0
        for p in self.params0:
            p.requires_grad = False

        # trainable tangent direction
        self.dparams = nn.ParameterList(
            [nn.Parameter(torch.zeros_like(p)) for p in params0]
        )

        # frozen forward
        self.func0 = lambda params, x: func(params, buffers0, x)

    def forward(self, x):
        # linearized update via JVP
        out, dp = jvp(self.func0, (self.params0,), (list(self.dparams),))
        return out + dp
\end{lstlisting}
\label{alg:linear}
\end{algorithm}

\vspace{-0.2cm}
\subsection{Compositional and Controllable Alignment}
\vspace{-0.2cm}
TS-DPO represents alignment objectives as directions in parameter space, rather than as scalarized rewards or task-specific fine-tuning. For each preference axis, such as helpfulness and verbosity, the method learns a separate tangent-space update:
\[
\tau_{\text{help}}, \; \tau_{\text{verb}} \in \mathbb{R}^{|\theta_0|},
\]
by two separate training. These updates play the role of basis vectors that characterize how the model parameters should change to improve along each preference dimension. The resulting aligned model is expressed through a linear combination of the base parameters and the learned preference directions: \looseness=-2
\[
\theta(\lambda)
=
\theta_{0}
+ \lambda_1 \tau_{\text{help}}
+ \lambda_2 \tau_{\text{verb}},
\]
where $(\lambda_1,\lambda_2)$ determine the strength and direction of alignment.
This represents multi-objective preference learning as a geometric operation in
tangent space. \looseness=-2

At inference time, controllable alignment is achieved simply by choosing $(\lambda_1,\lambda_2)$ and evaluating the model with the corresponding parameterization $\theta(\lambda)$. Thus, the preference coefficients directly specify the point in the helpfulness--verbosity trade-off space at inference time, without requiring retraining, additional reward functions, or separate models for each setting. This formulation allows the alignment behavior of the model to be manipulated as a continuous function of $(\lambda_1,\lambda_2)$, turning preference optimization into a geometry problem in parameter space. Different settings of $(\lambda_1,\lambda_2)$ produce a family of aligned models, enabling smooth interpolation or extrapolation across alignment axes while keeping the original checkpoint intact. \looseness=-2

\vspace{-0.2cm}
\section{Experiments and Results}
\label{sec:results}
\vspace{-0.2cm}
\subsection{Experimental Setup}
\vspace{-0.2cm}
We consider two alignment axes: helpfulness and verbosity. For helpfulness we use UltraFeedback \citep{ultrafeedback2023}, and for verbosity we use HelpSteer2 \citep{helpsteer2023}. From each dataset, we sample 6{,}000 preference pairs for training and reserve 2{,}000 examples for evaluation. Following practices of \citet{dpa2024}, we average instruction-following, truthfulness, honesty, and helpfulness ratings of UltraFeedback for the overall helpfulness objective. We use HelpSteer’s verbosity attribute for the verbosity objective. Throughout the experiments, we use \texttt{Llama-3.2-1B-Instruct}\footnote{\url{https://huggingface.co/meta-llama/Llama-3.2-1B-Instruct}} model. 

We train two separate models, one per preference axis, using standard DPO with chat-formatted pairs. The base model remains frozen in all settings. In TS-DPO, only the parameters corresponding to the last 16 transformer layers and the language model head are trainable, implemented through the tangent-space update directions $\tau_{\text{help}}$ and $\tau_{\text{verb}}$. These directions produce the aligned model at inference time via the compositional form $\theta(\lambda)$ from the previous section.

\paragraph{Baselines.}
We compare TS-DPO against two approaches. The first is scalarized DPO (DPO-Mixed), which optimizes a single combined objective following Eq.~(\ref{eq:dpo_loss}), using the combined helpfulness-verbosity dataset with 12{,}000 pairs. The second is Task-Vector DPO (directly denoted as DPO in the results), where two independent DPO models are trained on the two datasets with the same evaluation setup as TS-DPO, and their parameter deltas are composed as
\[
\theta_{\text{mix}} =
  \theta_{0}
  + \lambda_1\Delta\theta_{\text{help}}
  + \lambda_2\Delta\theta_{\text{verb}}.
\]

\paragraph{Reproducibility.}
All methods share the same training setup: one epoch, global batch size 32, gradient checkpointing (chunk size 32), and sequence length 1{,}024. We fine-tune on a single NVIDIA V100 (32GB) GPU on the Stanford Sherlock cluster. We use AdamW with the default HuggingFace \texttt{DPOTrainer} settings \texttt{(betas=(0.9,0.999), weight\_decay=0.1)}, mixed-precision \texttt{bfloat16}, and the DPO Bradley–Terry scale $\beta = 0.01$. Standard Task-Vector DPO takes $\approx$2.5 hours per objective; the DPO-Mixed baseline (twice the data) takes roughly double. TS-DPO incurs a JVP operation per forward pass, increasing runtime to $\approx$3.5 hours under the same configuration. When available, we also benchmarked on an NVIDIA H100-80GB, which reduces TS-DPO training to $\approx$15 minutes and supports larger effective batch sizes. \looseness=-2

For each preference direction, we sweep learning rates in the range $5\times10^{-6}$ to $5\times10^{-5}$ and select the best checkpoint based on validation performance on a held-out split (See Appx. \ref{app:desc} for details). \looseness=-2

\vspace{-0.2cm}
\subsection{Evaluation Metrics}
\vspace{-0.2cm}
We assess controllable alignment along the helpfulness and verbosity axes using two complementary evaluation pipelines: (i) preference comparison accuracy on held-out DPO datasets, and (ii) reward-model scoring on generated responses. The goal is to measure both the latent preference structure and the surface-level behavior as open-ended text generation induced by mixing preference directions.
\vspace{-0.2cm}
\paragraph{Pairwise Preference Accuracy.}
For each preference dataset $\mathcal{D}$, we evaluate whether the model assigns higher likelihood to the preferred response $y^+$ compared to $y^-$. Given a prompt $x$, the score is computed without generation:
\[
\operatorname{Acc}(f)
= \mathbb{P}[\log p_\theta(y^+\mid x) > \log p_\theta(y^-\mid x)].
\]
Likelihoods are obtained by running the model forward on the concatenated prompt–completion and averaging the token-level log-probability of the continuation. To test controllability, we sweep over mixtures of preference directions and recompute accuracy across datasets, forming the Pareto frontier. \looseness=-2
\vspace{-0.2cm}
\paragraph{Reward-Model Evaluation.}
We additionally measure performance using the frozen multi-objective reward model $\mathbf{r}_\phi(x,y)$ trained on HelpSteer and UltraFeedback (Mistral-7B-based; \texttt{RLHFlow/RewardModel-Mistral-7B-for-DPA-v1} \citep{dpa2024}), which scores each generated answer along multiple alignment dimensions. For every prompt, we generate a response using greedy decoding. and compute the RM score vector, reporting the helpfulness and verbosity dimentions $(r_\text{help}, r_\text{verb})$. We again sweep over the mixed direction $\Delta\theta_\alpha$ and compute the mean reward, producing a second Pareto frontier that reflects behavioral preferences during free-form generation. \looseness=-2

\vspace{-0.2cm}
\subsection{Preference Optimization Pareto Coverage}
\vspace{-0.2cm}
\begin{figure}[t!]
    \centering
    \subfloat[]{\includegraphics[width=0.45\textwidth]{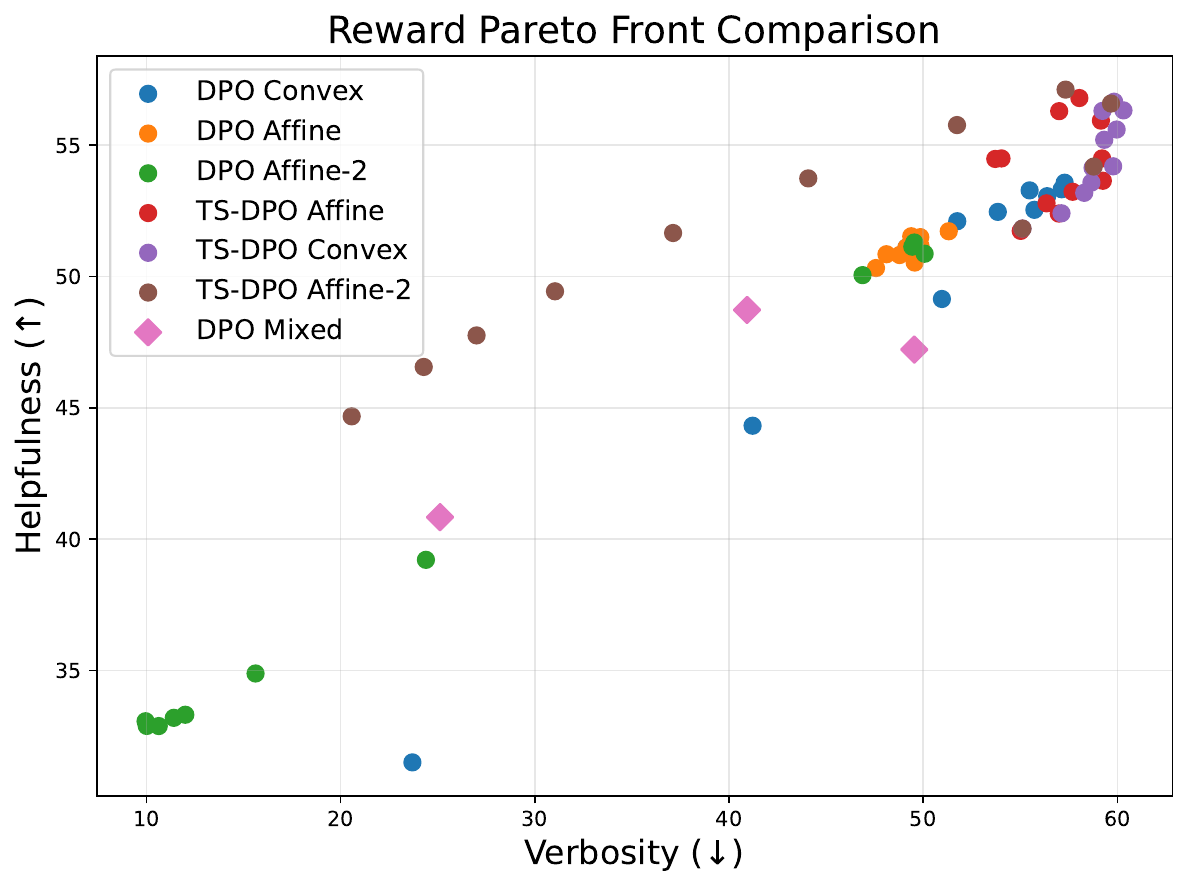}
    \label{fig:pareto2}} \quad
    \subfloat[]{\includegraphics[width=0.45\textwidth]{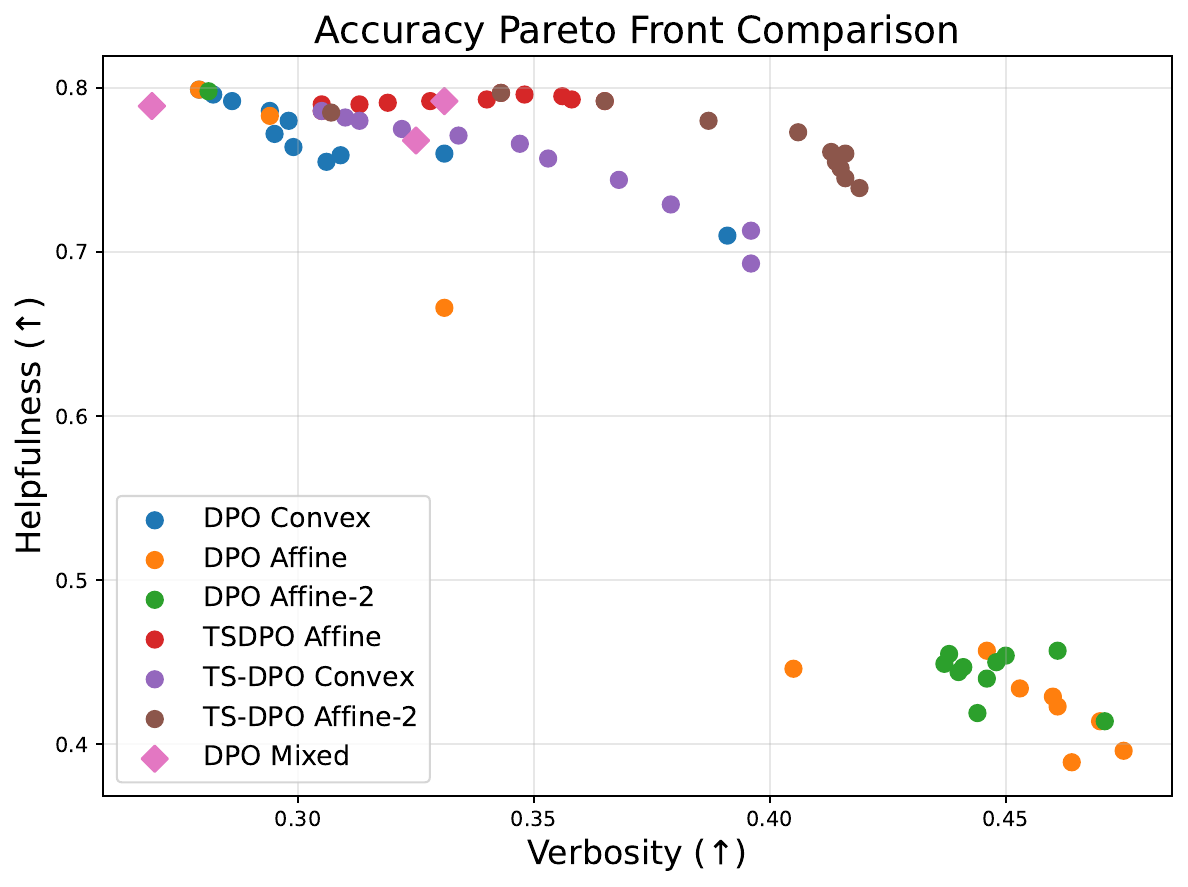}
    \label{fig:pareto1}}
    \caption{Pareto frontiers for controllable alignment along the helpfulness–verbosity axes. Each curve corresponds to a different sweep over the mixing coefficients $(\lambda_1,\lambda_2)$, including Convex, Affine, and Affine-2 constructions. The TS-DPO variants produce smoother frontiers, achieving higher helpfulness at comparable verbosity than Task-Vector DPO (DPO). The scalarized DPO baseline (DPO Mixed) is shown for three different points corresponding to various learning rates.}
    \label{fig:pareto}
    \vspace{-0.5cm}
\end{figure}

To evaluate controllable alignment across both preference axes, we sweep over the mixing coefficients $(\lambda_1,\lambda_2)$ corresponding to helpfulness and verbosity directions for all methods. We consider three types of mixtures: (i) Convex, where $\lambda\in\{0.0,0.1,\ldots,1.0\}$ and $(\lambda_1,\lambda_2)=(\lambda,1-\lambda)$; (ii) Affine, which allows asymmetric combinations and fixes the helpfulness direction, $(\lambda_1,\lambda_2)=(1,\lambda)$ with $\lambda\in\{0.0,0.1,\ldots,1.0\}$; and (iii) Affine-2, which extrapolates beyond the convex region, $(\lambda_1,\lambda_2)=(1,\lambda)$ with $\lambda\in\{0.0,0.5,\ldots,5.0\}$. The latter two are motivated by the empirical difficulty of improving the verbosity axis, so we vary that direction while holding the helpfulness update fixed. \looseness=-2

These three configurations correspond to the groups of plots in the experiments (\emph{DPO Convex}, \emph{DPO Affine}, and \emph{DPO Affine-2} for Task-Vector DPO), and their TS-DPO counterparts trained under the same strategy. We additionally include a scalarized baseline, \emph{DPO Mixed}, obtained by training on the concatenation of both datasets, for which we report results across multiple learning rates. \looseness=-2

Figure~\ref{fig:pareto2} shows the resulting Pareto fronts in reward space evaluated on 100 HelpSteer2 prompts. Across almost the entire frontier, the TS-DPO curves produce higher helpfulness for comparable or lower verbosity than their DPO counterparts. Increasing $\lambda$ moves the TS-DPO models smoothly along the frontier, which is consistent with additive and stable behavior in tangent space. In contrast, Task-Vector DPO and scalarized DPO exhibit non-linear behavior for the Affine-2 sweep: some settings collapse in helpfulness or move discontinuously, indicating instability under full-parameter composition. Although the Affine-2 variant of DPO reaches very low verbosity in some configurations, this comes at the cost of substantial drops in helpfulness. \looseness=-2

Figure~\ref{fig:pareto1} reports the same comparison using accuracy-based metrics. The three TS-DPO curves again cover the desirable part of the Pareto frontier, while scalarized and Task-Vector DPO fail to achieve combinations of low verbosity and high helpfulness. The Affine-2 sweep of Task-Vector DPO reaches a lower verbosity than TS-DPO, but only with very low helpfulness accuracy. A more extensive parameter search may explore this region further, but the current sweeps suggest that TS-DPO provides a more stable and favorable trade-off in practice. \looseness=-2

Overall, TS-DPO produces smooth and well-behaved Pareto frontiers, and dominates scalarized DPO across all hyperparameter configurations included in the plots. The improvements are especially pronounced when extrapolating beyond the convex mixture region, where tangent-space composition continues to behave predictably, while direct parameter composition does not. \looseness=-2

\vspace{-0.25cm}
\subsection{Analysis of Alignment Geometry}
\vspace{-0.2cm}
To better understand how TS-DPO organizes preference information, we analyze the geometry of the learned updates both in parameter space and in representation space.

\begin{figure}[t!]
    \centering
    \subfloat[\textbf{}]{
        \includegraphics[width=0.47\textwidth]{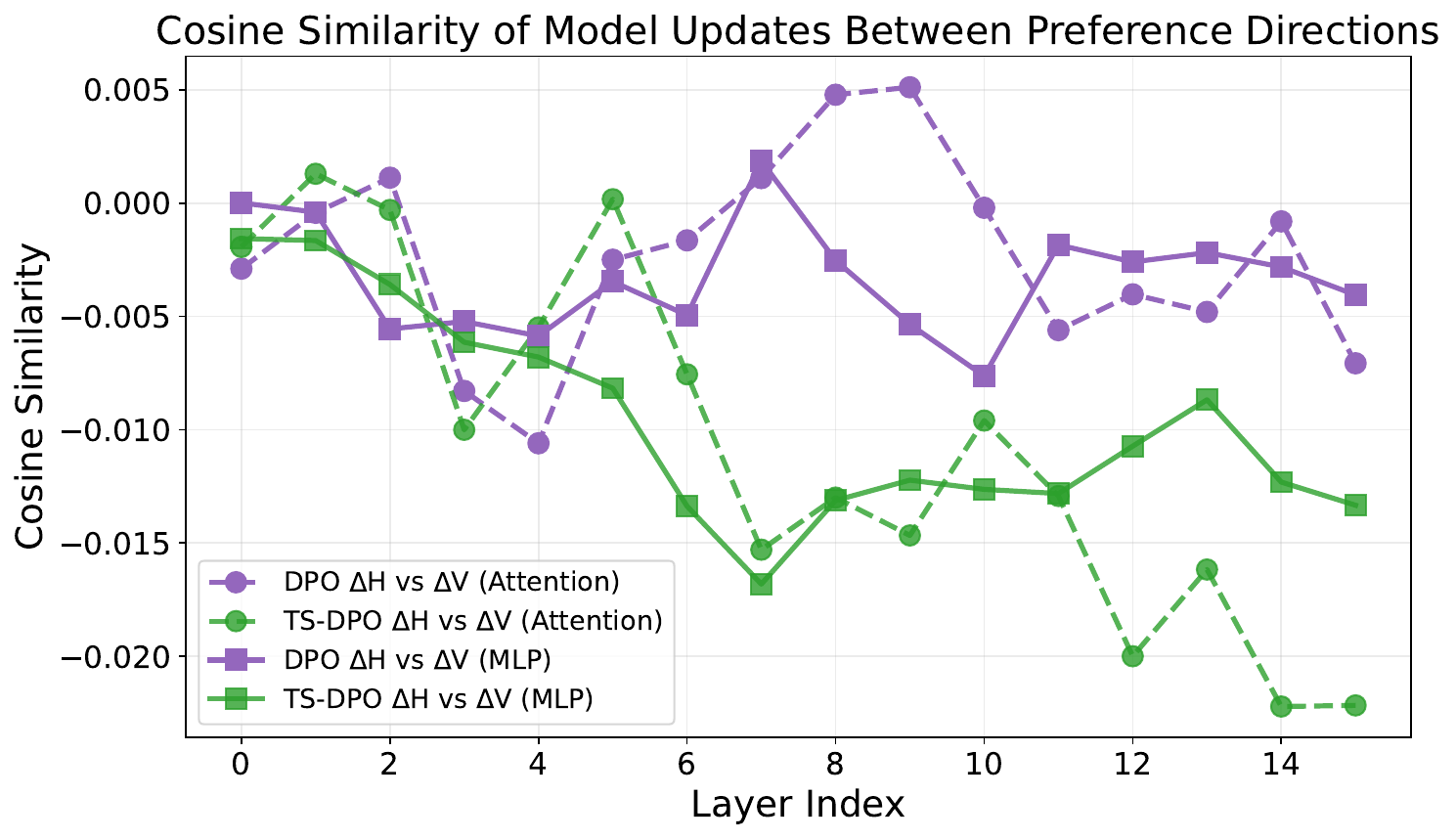} \label{fig:update_geometry1}
    } \quad
    \subfloat[\textbf{}]{
        \includegraphics[width=0.463\textwidth]{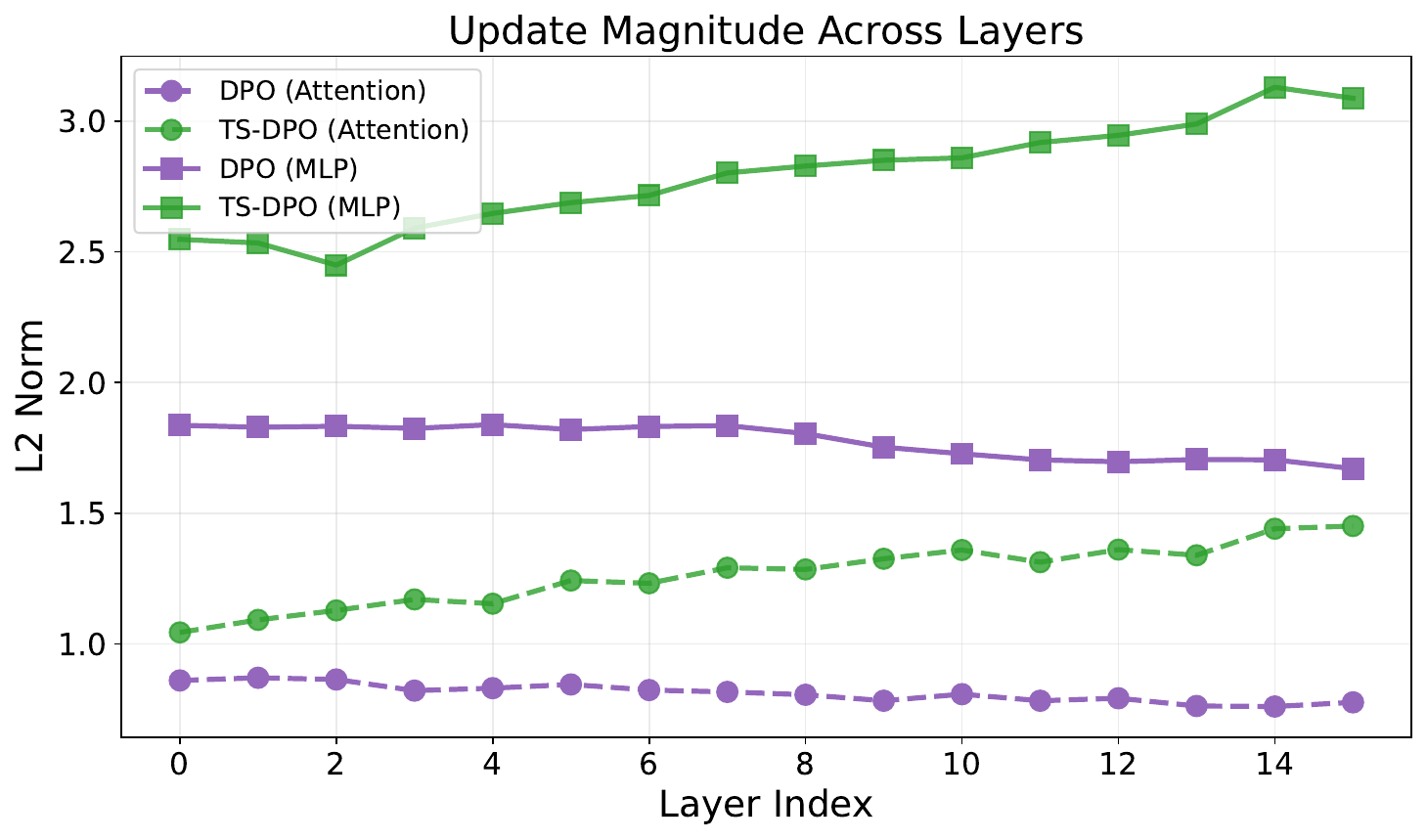} \label{fig:update_geometry2}
    }
    \caption{Layerwise geometry of preference updates for DPO and TS-DPO. We decompose parameter deltas into attention and MLP blocks and measure (a) cosine similarity between helpfulness and verbosity directions and (b) $\ell_2$-norm of helpfulness updates across layers.}
    \label{fig:update_geometry} 
    \vspace{-0.5cm}
\end{figure}

In Fig.~\ref{fig:update_geometry1}, we measure the cosine similarity between the helpfulness and verbosity update directions on a per-layer basis, separating attention and MLP components. For each transformer layer $L$, we extract the parameter deltas $\Delta\theta_{\text{help}}^{(L)}$ and $\Delta\theta_{\text{verb}}^{(L)}$ relative to the frozen base model and compute their cosine similarity. Across both DPO and TS-DPO, the similarities remain close to zero in most layers, indicating that the two objectives are implemented via nearly orthogonal adjustments in weight space. Notably, TS-DPO consistently achieves even lower similarity, implying a cleaner separation between the helpfulness and verbosity axes. This suggests that TS-DPO preserves more disentangled preference directions rather than blending various objectives. \looseness=-2

Fig.~\ref{fig:update_geometry2} shows the $\ell_{2}$-norm of the per-layer updates. For both methods, MLP blocks carry larger updates than attention blocks, suggesting that preference adaptation primarily occurs in the feed-forward components. TS-DPO concentrates its updates in deeper layers—those most associated with task-specific behavior, whereas standard DPO distributes updates more uniformly. Although TS-DPO produces larger norms overall, these updates live purely in the tangent space and do not directly modify the base parameters, so their effect is qualitatively different from full fine-tuning. \looseness=-2

We next analyze alignment in \emph{activation space}. For a shared set of prompts, we record last-token hidden states and form functional deltas relative to base model:
\[
\Delta h_{\text{help}}(x)=h_{\text{help}}(x)-h_{\text{base}}(x),\qquad
\Delta h_{\text{verb}}(x)=h_{\text{verb}}(x)-h_{\text{base}}(x).
\]
\begin{wrapfigure}{r}{0.45\textwidth}
\centering
    \centering
    \vspace{-0.15cm}
    \includegraphics[width=0.38\textwidth]{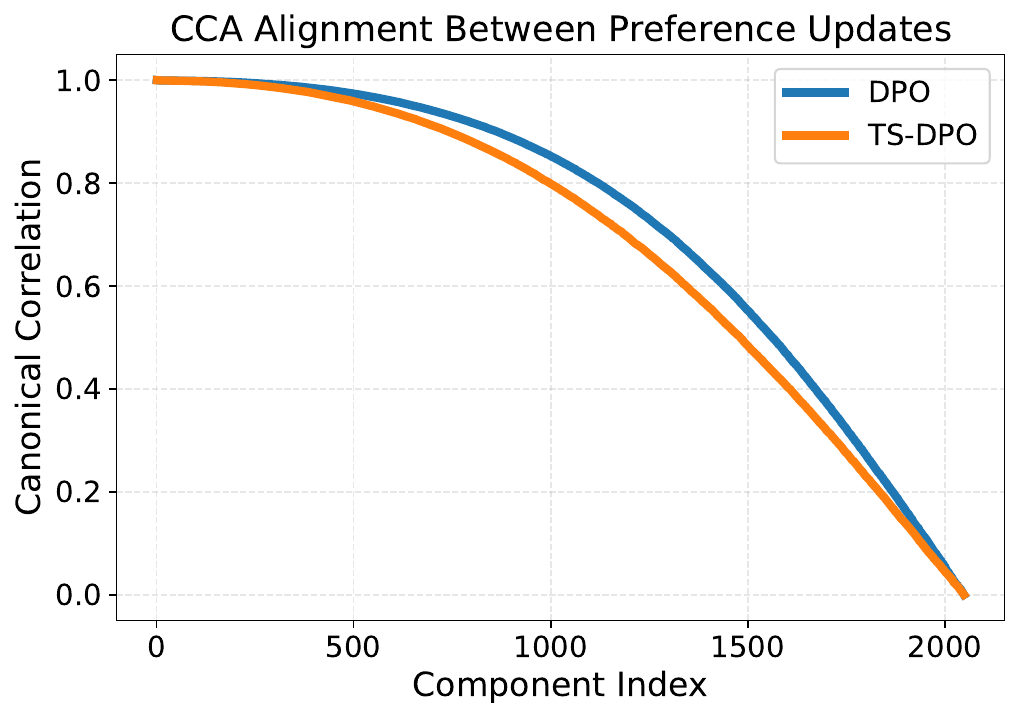}
    \caption{Canonical correlation spectrum between helpfulness and verbosity activation deltas for DPO and TS-DPO.}
    \label{fig:cca_alignment}
    \vspace{-0.7cm}
\end{wrapfigure}
For TS-DPO, these are obtained via the JVP linearization ($\Delta f_i(x)=J_{\theta_0}(x)(\theta_i-\theta_0)$) which gives the first-order effect of fine-tuning while keeping the base weights fixed. We then apply canonical correlation analysis (CCA) between the two sets of deltas. As shown in Fig.~\ref{fig:cca_alignment}, the TS-DPO spectrum exhibits a faster decay than DPO, meaning that the variance remains in a more constrained subspace, indicating sharper and more disentangled functional separation of preferences. \looseness=-2

These results provide complementary evidence that TS-DPO organizes preference updates into distinct and compositional directions.  \looseness=-2

\subsection{Further Discussions}

TS-DPO reliably learned separate preference directions without updating the base parameters. The method produced smooth interpolation along each axis and, in most settings, achieved broader Pareto coverage than standard DPO. The learned update directions were interpretable and aligned well with the intended preferences, enabling controllable shifts in helpfulness and verbosity without retraining the base model. \looseness=-2

We were not able to evaluate DPA \citep{dpa2024} end-to-end because its full training code is not publicly released, limiting direct comparison with the closest related method. Furthermore, in Figure \ref{fig:pareto}, TS-DPO also did not recover the entire Pareto front---some trade-off regions such as the ones giving the best verbosity, remain difficult to reach. The mixing strategy requires further tuning, and additional experiments are needed to fully establish consistent improvements over task-vector DPO. \looseness=-2

Linearizing alignment in the tangent space is practical and competitive. The approach yields modular preference directions, and improves controllability. The results suggest that multi-objective alignment benefits from representing each preference as an independent geometric direction in parameter space rather than collapsing them into a scalar objective. \looseness=-2
\vspace{-0.3cm}
\paragraph{Limitations.}
TS-DPO introduces additional computational overhead because each forward pass requires a Jacobian–vector product. As a result, training is slower than standard DPO, and our current implementation does not yet exploit KV caching. For preference-conditioned decoding, we relied on greedy generation in the reward model evaluation pipeline due to the lack of full support for alternative decoding strategies; this restricts the expressiveness of the evaluation and may bias results toward shorter responses. Moreover, we were unable to include DPA as a baseline because the full training implementation is not publicly available, limiting comparison against the closest multi-objective preference alignment method. Although our experiments demonstrate that TS-DPO improves controllability and Pareto coverage, further evaluation is needed on more datasets, larger architectures, and more than two preferences before making definitive claims about scalability and superiority. \looseness=-2
\vspace{-0.2cm}

\section{Conclusion and Future Work}
\vspace{-0.2cm}
This work introduced Tangent-Space Direct Preference Optimization (TS-DPO) as a method for learning disentangled preference directions. Instead of updating the base model weights, TS-DPO learns additive tangent directions in the parameter space of the pre-trained model. Our experiments on helpfulness and verbosity demonstrate that TS-DPO produces modular and controllable preference updates, achieves improved Pareto coverage compared to standard DPO, and supports smooth interpolation between trade-off points without retraining the base model. We also provided an analysis of representational geometry using CCA and subspace alignment.
\vspace{-0.25cm}
\paragraph{Implications.}
TS-DPO offers a promising mechanism for value-sensitive, context-aware deployment: aligning the same model to different user or system constraints becomes a composition problem rather than a re-training problem. This view brings preference optimization closer to modular control, where alignment is not a single solution but a family of controllable behaviors.
\vspace{-0.25cm}
\paragraph{Future Work.}
We see several natural extensions. First, scaling TS-DPO to more than two objectives (e.g., helpfulness–safety–conciseness) would test whether its tangent-space basis remains disentangled in higher dimensions. Second, improved vector mixing or normalization schemes may offer broader Pareto coverage. Third, evaluating TS-DPO on larger models and datasets would clarify whether its gains persist at scale. Finally, once public implementations are available, TS-DPO should be compared with modular alignment methods such as DPA and model-merging approaches.

Overall, our findings support the idea that linearizing around the base model is a practical and principled approach for multi-objective alignment, and raise the possibility that alignment itself can be decomposed into stable, modular update directions rather than monolithic end-to-end training. This viewpoint may help move toward AI systems that are both more adaptable and more interpretable.

\vspace{-0.2cm}
\section*{Statement on Plagiarism, Bias, and Inaccuracies} 
\vspace{-0.2cm}
In this report, we ensured academic integrity by carefully attributing all conceptual foundations, including DPO and tangent space training, to their respective authors. We avoided textual plagiarism by paraphrasing ideas in our own words and cross-checking against original sources to prevent unintentional overlap. To mitigate non-attribution of ideas, we cited prior work whenever their assumptions, formulations, or empirical observations informing our methodology. We addressed potential inaccuracies by validating all experimental claims. To reduce bias, we used  standardized datasets (UltraFeedback, HelpSteer2) and reward models rather than subjective judgments. Throughout, we adhered to Stanford's expectations of academic integrity by maintaining  transparent citations, careful interpretation, and an honest presentation.

\vspace{-0.2cm}
\section*{Reflection on Use of AI Tools}
\vspace{-0.2cm}
AI tools are used primarily as writing and debugging assistants rather than generators of technical content. During the project, we occasionally consulted models to clarify PyTorch usage, linearization mechanics, and evaluation protocols, but all methodological decisions such mixture strategies were developed independently and derived directly from the literature. When drafting the report, we used AI tools for stylistic refinement, consistency checking, and identifying unclear explanations, which improved readability without altering the underlying scientific contributions. This experience reinforced that AI assistance is valuable for accelerating iteration and detecting ambiguities. Going forward, we expect to use AI similarly as an aid for editing, code diagnostics, and literature navigation, while ensuring that conceptual, mathematical, and empirical work remains original.

\vspace{-0.2cm}
\section*{Impact Statement}
\vspace{-0.2cm}
This work investigates controllable multi-objective preference alignment in LLMs through tangent-space updates, which offers both societal benefits and potential risks. Benefits include enabling adaptive communication styles and facilitating value-sensitive deployment by treating alignment as a compositional control. But, the same mechanisms could be misused to amplify undesirable behaviors. For example, adjusting verbosity or helpfulness to obscure harmful intent or manipulate users. To mitigate such concerns, we used only public datasets (UltraFeedback, HelpSteer2), avoided sensitive data, and reported limitations transparently. Although no human subjects were involved, the project aligns with Stanford’s standards for responsible development by emphasizing transparency, reproducibility, and careful consideration of the ethical implications of preference control.

\newpage

\bibliographystyle{apalike} %
\bibliography{references}

\newpage
\begin{center}

\vspace{1.5cm}
{\Large \fontseries{bx}\selectfont Appendix}
\end{center}

\appendix

\vspace{0.5cm}
\section{Further Details of Pareto Front Experiments }
\label{app:desc}

Table \ref{tab:last} report the exact numerical values underlying the Pareto front plots in Sec.~\ref{sec:results}. For each method and mixture strategy (Convex, Affine, and Affine-2), we record the helpfulness and verbosity performance across a sweep of mixture coefficients $\lambda$. The first table corresponds to accuracy-based preference comparison using the original preference datasets. The second table reports the reward-model-based evaluation scores using the Mistral-7B DPA reward model.

For each method, rows enumerate different configurations of $(\lambda_1,\lambda_2)$, and columns list the resulting helpfulness and verbosity scores. These values match the scatter points in Figures~\ref{fig:pareto1} and~\ref{fig:pareto2}, and allow for precise comparison of the Pareto behavior rather than relying solely on visual inspection.

The Convex and Affine settings explore the trade-off between helpfulness and verbosity along constrained mixtures, while Affine-2 evaluates extrapolation beyond the convex region. TS-DPO generally provides broader coverage and smoother trade-offs across both evaluation metrics, whereas DPO Mixed produces a single operating point per training.

\begin{table}[t!]
\tiny
\centering
\caption{\small Combined Pareto results showing Helpfulness–Verbosity (H-V) alignment across both accuracy and reward metrics.}
\begin{tabular}{lcccccc}
\toprule
Method & $(\lambda_1,\lambda_2)$ & LR (H,V) & Acc-H↑ & Acc-V↑ & R-H↑ & R-V↓ \\
\midrule
\multicolumn{7}{l}{\textbf{DPO Convex}}\\
 & (0.0,1.0) & (5e-5,1e-5) & 0.710 & 0.391 & 51.11 & 49.14 \\
 & (0.1,0.9) & (5e-5,1e-5) & 0.760 & 0.331 & 52.11 & 51.75 \\
 & (0.2,0.8) & (5e-5,1e-5) & 0.759 & 0.309 & 52.46 & 53.84 \\
 & (0.3,0.7) & (5e-5,1e-5) & 0.755 & 0.306 & 53.32 & 57.13 \\
 & (0.4,0.6) & (5e-5,1e-5) & 0.764 & 0.299 & 53.28 & 55.48 \\
 & (0.5,0.5) & (5e-5,1e-5) & 0.772 & 0.295 & 53.06 & 56.38 \\
 & (0.6,0.4) & (5e-5,1e-5) & 0.780 & 0.298 & 53.57 & 57.28 \\
 & (0.7,0.3) & (5e-5,1e-5) & 0.786 & 0.294 & 52.54 & 55.72 \\
 & (0.8,0.2) & (5e-5,1e-5) & 0.792 & 0.286 & 49.15 & 50.96 \\
 & (0.9,0.1) & (5e-5,1e-5) & 0.796 & 0.282 & 44.32 & 41.21 \\
 & (1.0,0.0) & (5e-5,1e-5) & 0.799 & 0.279 & 31.50 & 23.69 \\
\midrule
\multicolumn{7}{l}{\textbf{DPO Affine}}\\
 & (1,0.0) & (5e-5,1e-5) & 0.799 & 0.279 & 50.54 & 49.56 \\
 & (1,0.1) & (5e-5,1e-5) & 0.783 & 0.294 & 50.82 & 48.78 \\
 & (1,0.2) & (5e-5,1e-5) & 0.666 & 0.331 & 50.33 & 47.57 \\
 & (1,0.3) & (5e-5,1e-5) & 0.446 & 0.405 & 50.85 & 48.10 \\
 & (1,0.4) & (5e-5,1e-5) & 0.389 & 0.464 & 51.26 & 49.77 \\
 & (1,0.5) & (5e-5,1e-5) & 0.396 & 0.475 & 51.72 & 51.31 \\
 & (1,0.6) & (5e-5,1e-5) & 0.414 & 0.470 & 51.22 & 49.63 \\
 & (1,0.7) & (5e-5,1e-5) & 0.429 & 0.460 & 51.50 & 49.85 \\
 & (1,0.8) & (5e-5,1e-5) & 0.434 & 0.453 & 51.17 & 49.86 \\
 & (1,0.9) & (5e-5,1e-5) & 0.457 & 0.446 & 51.53 & 49.38 \\
 & (1,1.0) & (5e-5,1e-5) & 0.423 & 0.461 & 51.11 & 49.14 \\
\midrule
\multicolumn{7}{l}{\textbf{DPO Affine-2}}\\
 & (1,0.0) & (5e-5,1e-5) & 0.798 & 0.281 & 51.30 & 49.54 \\
 & (1,0.5) & (5e-5,1e-5) & 0.414 & 0.471 & 51.14 & 49.43 \\
 & (1,1.0) & (5e-5,1e-5) & 0.419 & 0.444 & 50.87 & 50.07 \\
 & (1,1.5) & (5e-5,1e-5) & 0.450 & 0.448 & 50.06 & 46.87 \\
 & (1,2.0) & (5e-5,1e-5) & 0.440 & 0.446 & 39.21 & 24.39 \\
 & (1,2.5) & (5e-5,1e-5) & 0.457 & 0.461 & 34.89 & 15.62 \\
 & (1,3.0) & (5e-5,1e-5) & 0.447 & 0.441 & 33.32 & 12.00 \\
 & (1,3.5) & (5e-5,1e-5) & 0.455 & 0.438 & 33.20 & 11.41 \\
 & (1,4.0) & (5e-5,1e-5) & 0.454 & 0.450 & 32.89 & 10.64 \\
 & (1,4.5) & (5e-5,1e-5) & 0.444 & 0.440 & 32.89 & 10.01 \\
 & (1,5.0) & (5e-5,1e-5) & 0.449 & 0.437 & 33.07 & 9.95 \\
\midrule
\multicolumn{7}{l}{\textbf{TS-DPO Convex}}\\
 & (0.0,1.0) & (1e-5,5e-6) & 0.693 & 0.396 & 56.66 & 59.83 \\
 & (0.1,0.9) & (1e-5,5e-6) & 0.713 & 0.396 & 56.31 & 59.23 \\
 & (0.2,0.8) & (1e-5,5e-6) & 0.729 & 0.379 & 56.33 & 60.31 \\
 & (0.3,0.7) & (1e-5,5e-6) & 0.744 & 0.368 & 55.60 & 59.96 \\
 & (0.4,0.6) & (1e-5,5e-6) & 0.757 & 0.353 & 55.21 & 59.32 \\
 & (0.5,0.5) & (1e-5,5e-6) & 0.766 & 0.347 & 54.14 & 58.72 \\
 & (0.6,0.4) & (1e-5,5e-6) & 0.771 & 0.334 & 54.20 & 59.78 \\
 & (0.7,0.3) & (1e-5,5e-6) & 0.775 & 0.322 & 53.58 & 58.66 \\
 & (0.8,0.2) & (1e-5,5e-6) & 0.780 & 0.313 & 53.19 & 58.29 \\
 & (0.9,0.1) & (1e-5,5e-6) & 0.782 & 0.310 & 52.41 & 57.12 \\
 & (1.0,0.0) & (1e-5,5e-6) & 0.786 & 0.305 & 51.83 & 55.11 \\
\midrule
\multicolumn{7}{l}{\textbf{TS-DPO Affine}}\\
 & (1,0.0) & (1e-5,5e-6) & 0.785 & 0.305 & 54.50 & 54.03 \\
 & (1,0.1) & (1e-5,5e-6) & 0.790 & 0.305 & 54.48 & 53.71 \\
 & (1,0.2) & (1e-5,5e-6) & 0.790 & 0.313 & 56.30 & 57.00 \\
 & (1,0.3) & (1e-5,5e-6) & 0.791 & 0.319 & 56.80 & 58.04 \\
 & (1,0.4) & (1e-5,5e-6) & 0.792 & 0.328 & 55.94 & 59.15 \\
 & (1,0.5) & (1e-5,5e-6) & 0.793 & 0.340 & 54.50 & 59.21 \\
 & (1,0.6) & (1e-5,5e-6) & 0.797 & 0.343 & 53.65 & 59.25 \\
 & (1,0.7) & (1e-5,5e-6) & 0.796 & 0.348 & 53.23 & 57.69 \\
 & (1,0.8) & (1e-5,5e-6) & 0.795 & 0.356 & 52.79 & 56.35 \\
 & (1,0.9) & (1e-5,5e-6) & 0.793 & 0.358 & 52.40 & 56.98 \\
 & (1,1.0) & (1e-5,5e-6) & 0.792 & 0.365 & 51.74 & 55.02 \\
\midrule
\multicolumn{7}{l}{\textbf{TS-DPO Affine-2}}\\
 & (1,0.0) & (1e-5,5e-6) & 0.785 & 0.307 & 51.83 & 55.11 \\
 & (1,0.5) & (1e-5,5e-6) & 0.797 & 0.343 & 54.19 & 58.78 \\
 & (1,1.0) & (1e-5,5e-6) & 0.792 & 0.365 & 56.59 & 59.67 \\
 & (1,1.5) & (1e-5,5e-6) & 0.780 & 0.387 & 57.12 & 57.33 \\
 & (1,2.0) & (1e-5,5e-6) & 0.773 & 0.406 & 55.77 & 51.74 \\
 & (1,2.5) & (1e-5,5e-6) & 0.761 & 0.413 & 53.74 & 44.08 \\
 & (1,3.0) & (1e-5,5e-6) & 0.760 & 0.416 & 51.66 & 37.12 \\
 & (1,3.5) & (1e-5,5e-6) & 0.755 & 0.414 & 49.44 & 31.04 \\
 & (1,4.0) & (1e-5,5e-6) & 0.751 & 0.415 & 47.76 & 27.00 \\
 & (1,4.5) & (1e-5,5e-6) & 0.745 & 0.416 & 46.56 & 24.28 \\
 & (1,5.0) & (1e-5,5e-6) & 0.739 & 0.419 & 44.68 & 20.57 \\
\midrule
\multicolumn{7}{l}{\textbf{DPO Mixed}}\\
 & -- & (1e-5) & 0.768 & 0.325 & 47.22 & 49.54 \\
 & -- & (2.5e-5) & 0.792 & 0.331 & 48.73 & 40.93 \\
 & -- & (5e-5) & 0.789 & 0.269 & 40.84 & 25.12 \\
\bottomrule
\end{tabular}
\label{tab:last}
\end{table}

\newpage
\section{ Loss Curves From Learning Rate Tuning Experiments }
\label{app:param}

In this section, we present the loss curves from the learning rate tuning experiments for each training setting.

\begin{figure}[h!]
    \subfloat[\textbf{}]{
        \includegraphics[width=0.48\textwidth]{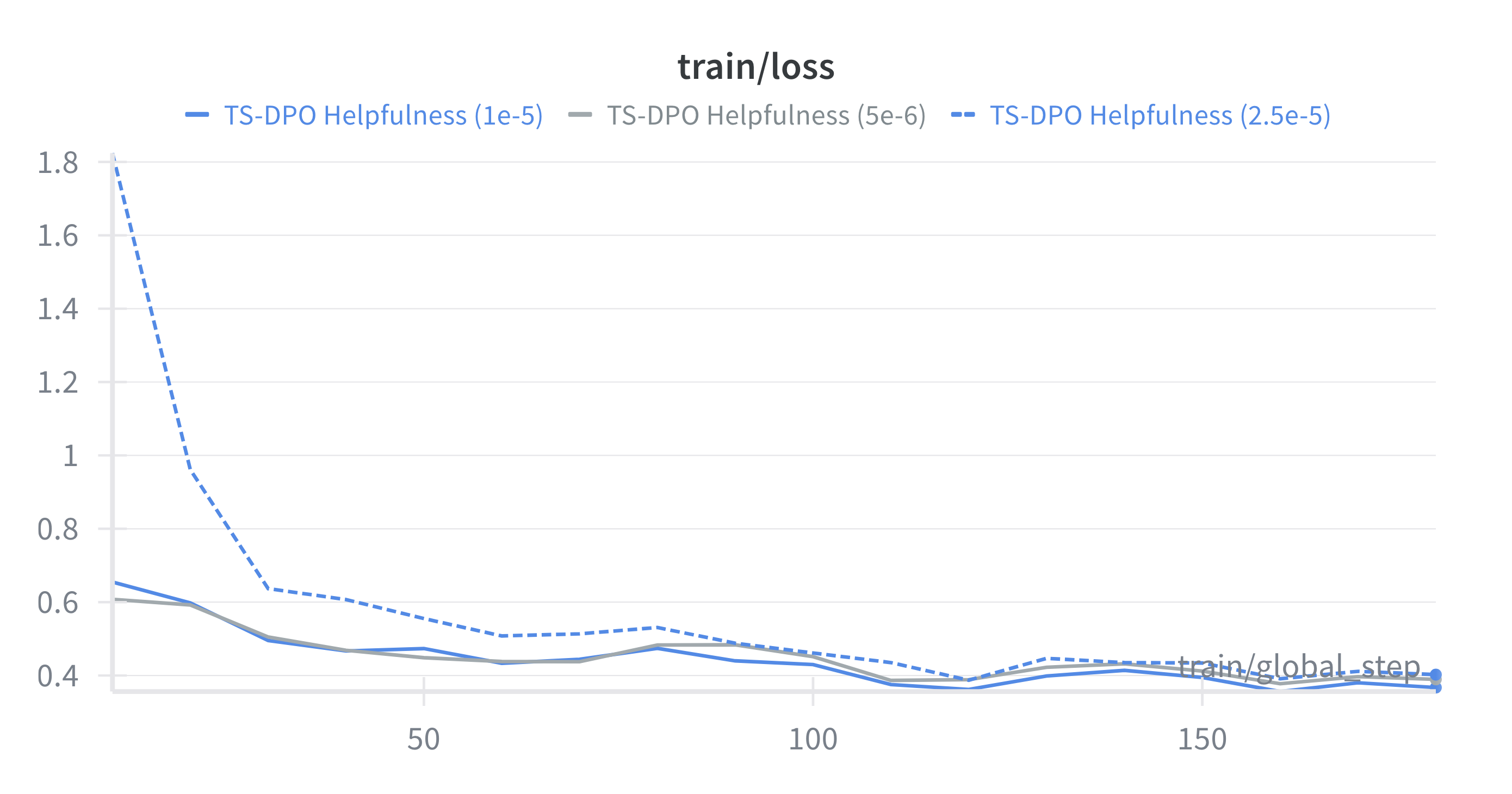} \label{fig:xx1}
    } \quad
    \subfloat[\textbf{}]{
        \includegraphics[width=0.48\textwidth]{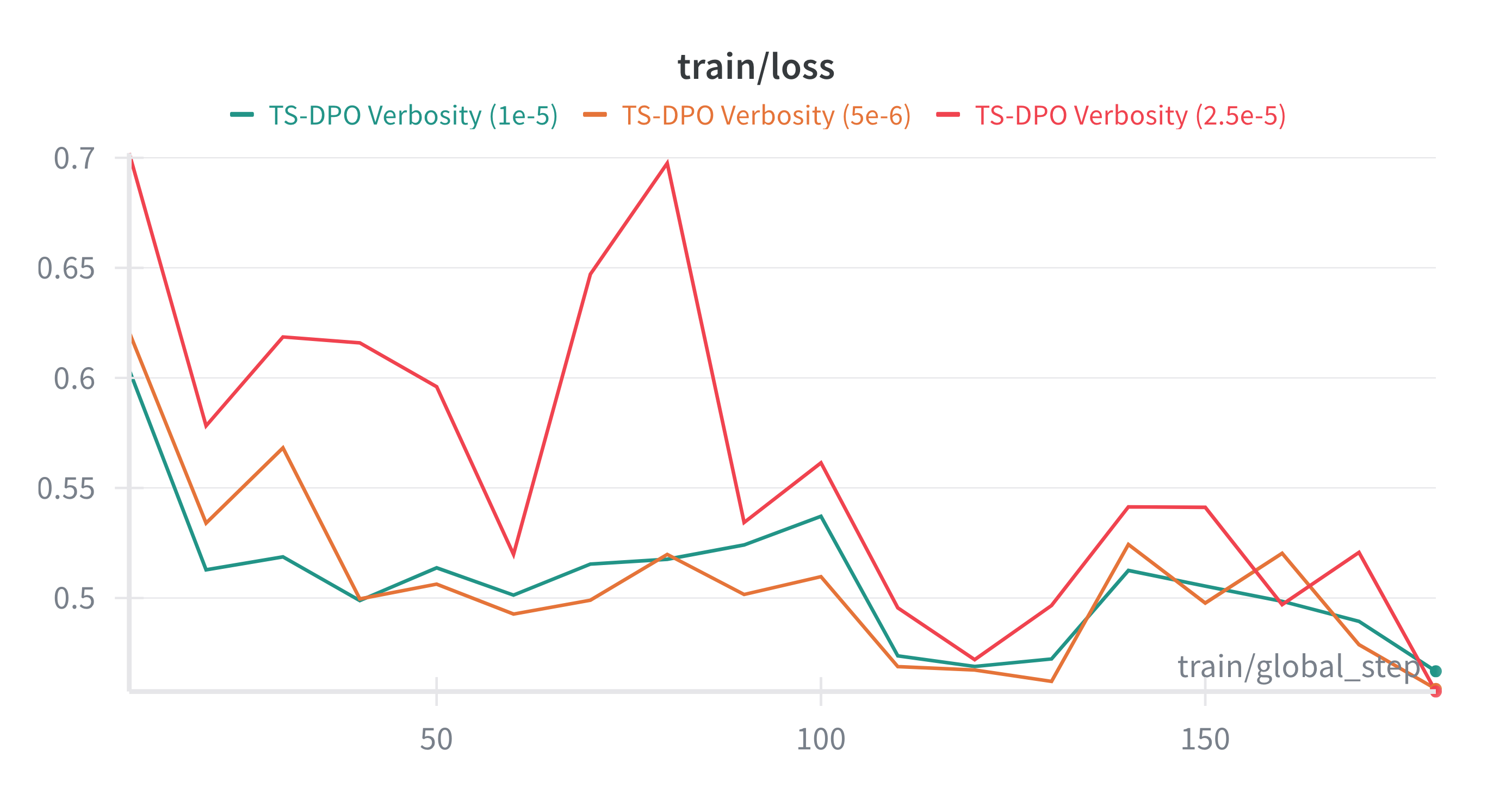} \label{fig:xx2}
    }
    \caption{TS-DPO tranining losses for various learning rates (a) helpfulness (b) verbosity alignment.}
    \label{fig:aaaaaaaa} 
\end{figure}

\begin{figure}[h!]
    \subfloat[\textbf{}]{
        \includegraphics[width=0.48\textwidth]{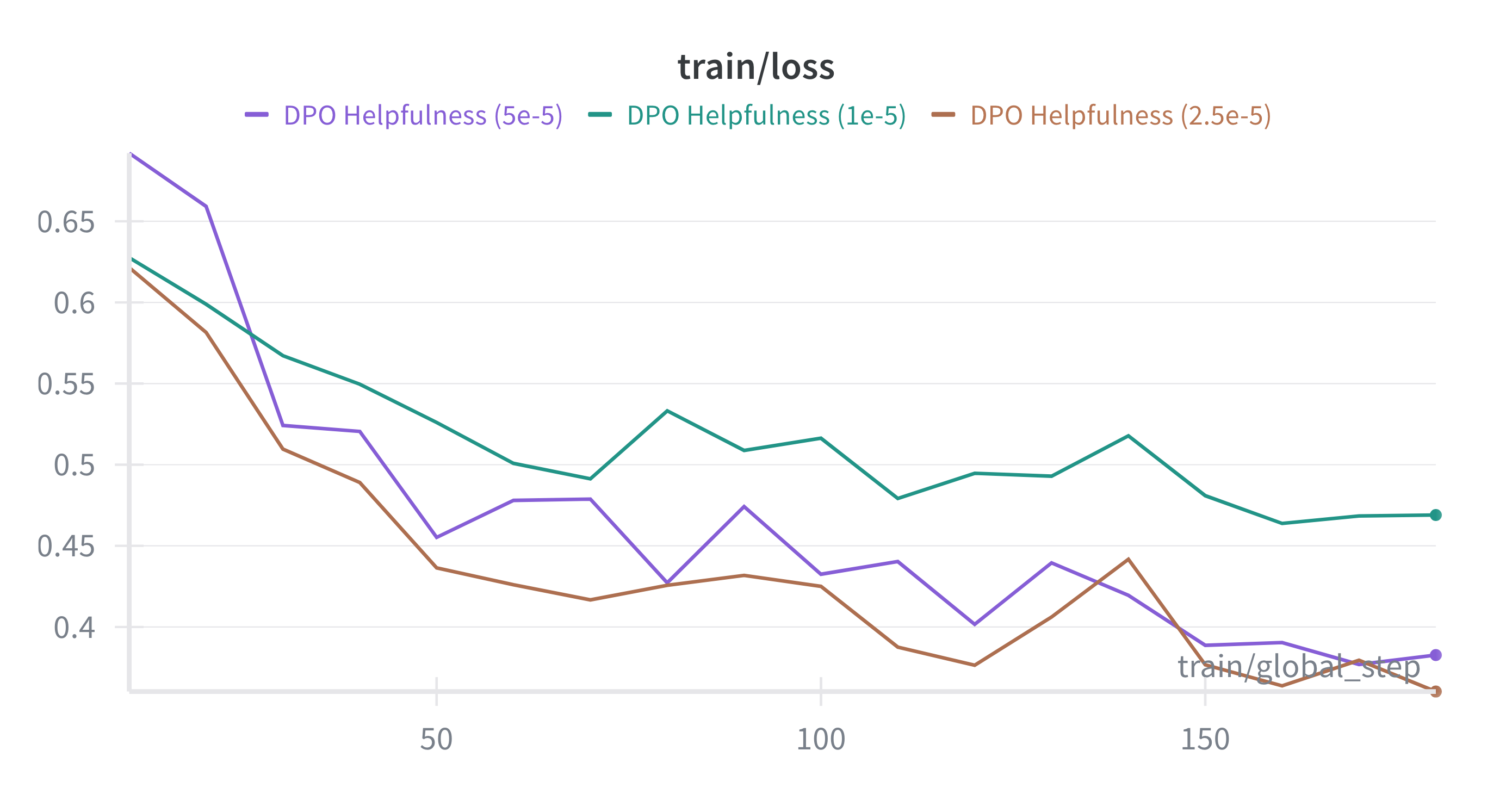} \label{fig:xx11}
    } \quad
    \subfloat[\textbf{}]{
        \includegraphics[width=0.48\textwidth]{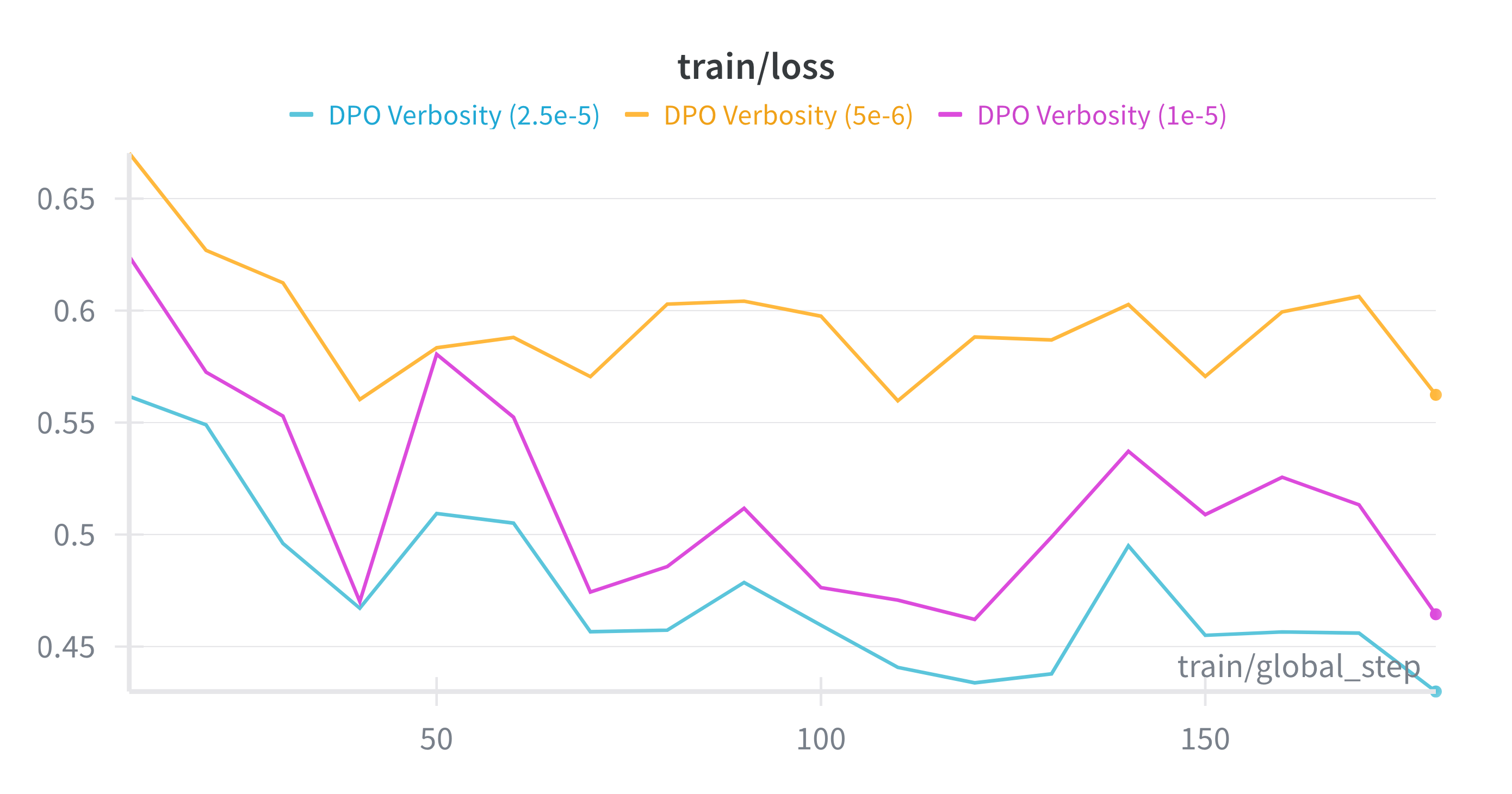} \label{fig:xx22}
    }
    \caption{Task Vector DPO tranining losses for various learning rates for (a) helpfulness (b) verbosity alignment.}
    \label{fig:aaaaaa} 
\end{figure}

\begin{figure}[h!]
\centering
    \subfloat[\textbf{}]{
        \includegraphics[width=0.48\textwidth]{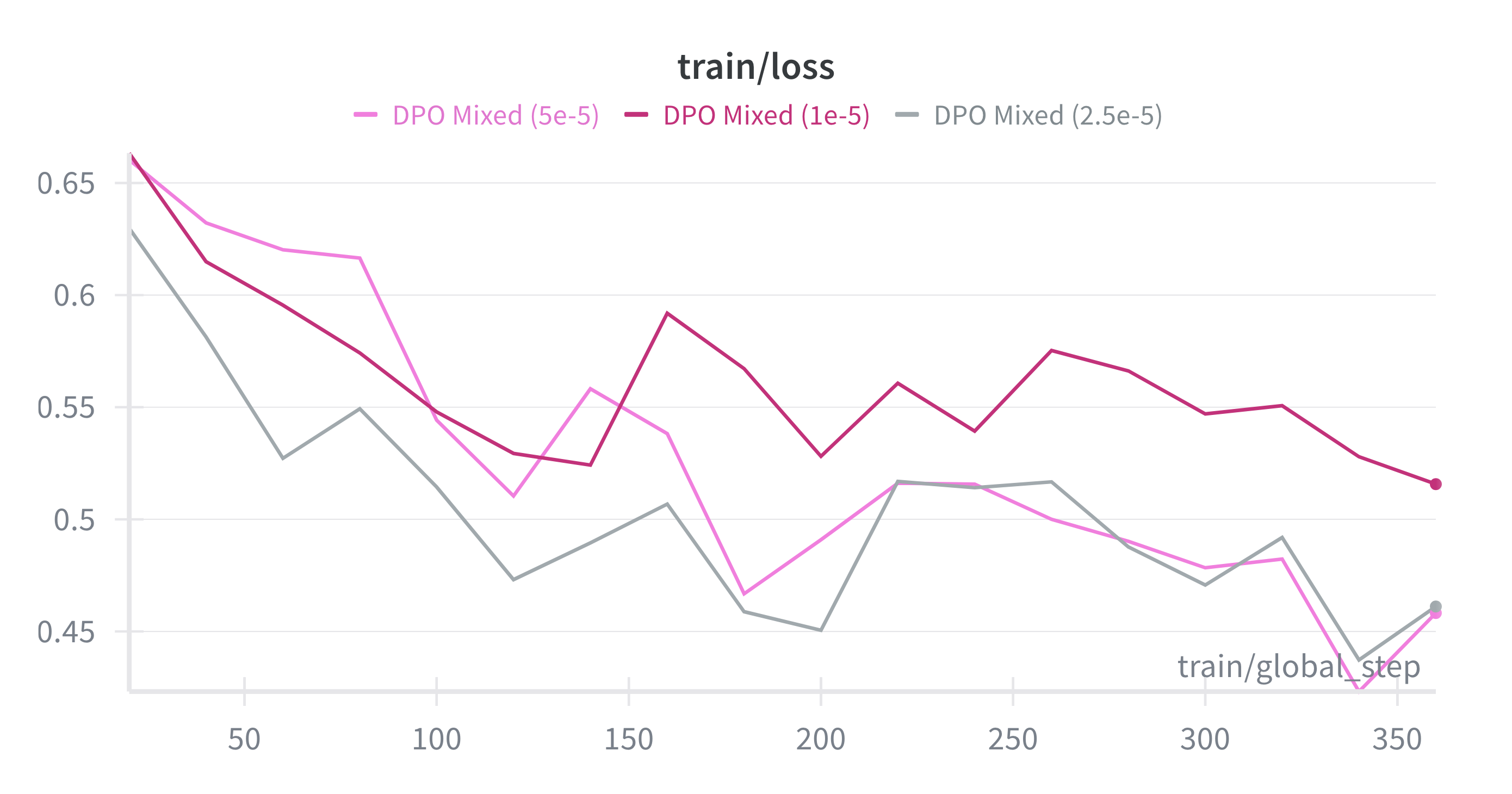} \label{fig:xx111}
    } 
    \caption{DPO Mixed tranining losses for various learning rates.}
    \label{fig:aaaa} 
\end{figure}

\end{document}